\pdfoutput=1

\documentclass[11pt]{article}

\usepackage[final]{acl}

\usepackage{times}
\usepackage{latexsym}

\usepackage[T1]{fontenc}

\usepackage[utf8]{inputenc}

\usepackage{microtype}

\usepackage{inconsolata}

\usepackage{enumitem}
\usepackage{rotating}

\usepackage{booktabs}
\usepackage{multirow}

\usepackage{xcolor,colortbl}

\definecolor{Gray}{gray}{0.94}
\definecolor{LightCyan}{rgb}{0.88,1,1}
\newcolumntype{a}{>{\columncolor{Gray}}c}
\newcolumntype{o}{>{\columncolor{white}}c}
\usepackage{soul}
\definecolor{celeste}{cmyk}{0.3922, 0.0353, 0, 0.1}
\definecolor{purple}{cmyk}{0.16, 0.28, 0, 0}
\definecolor{brilliantlavender}{cmyk}{0, 0.2235, 0, 0.1}
\definecolor{LightRed}{RGB}{232, 56, 107} 
\definecolor{LightBlue}{RGB}{116, 232, 226}
\definecolor{Tan}{rgb}{0.8203,0.7031,0.5469}
\definecolor{gblue}{RGB}{81,231,195}

\usepackage{tcolorbox}

\usepackage{tikz}
\usepackage{array}
\usetikzlibrary{decorations.pathreplacing}
\newcommand{\roundedtable}[2]{
    \begin{tikzpicture}
        \node (table) [
            rectangle, 
            rounded corners, 
            draw=black,
            inner sep=0pt
        ] {
            \begin{tabular}{#1}
                #2
            \end{tabular}
        };
    \end{tikzpicture}
}

\usepackage{amsthm}
\usepackage{amsmath}
\usepackage{amssymb}
\usepackage{xspace}
\usepackage{pifont}
\newcommand{\narrowtextsc}[1]{\textls[-50]{\textsc{#1}}}
\newcommand{\lm}[1]{\texttt{#1}}
\newcommand{\sys}[1]{\narrowtextsc{#1}}
\newcommand{\data}[1]{\textsf{#1}}
\newcommand{\present}{$\blacksquare$}

\newcommand\nnfootnote[1]{%
  \begin{NoHyper}
  \renewcommand\thefootnote{}\footnote{#1}%
  \addtocounter{footnote}{-1}%
  \end{NoHyper}
}

\title{CoXQL: A Dataset for Parsing Explanation Requests \\in Conversational XAI Systems}

\newcommand{\affilsup}[1]{\rlap{\textsuperscript{\normalfont#1}}}

\author{
    Qianli Wang\affilsup{1,2}
    \qquad 
    Tatiana Anikina$^*$\affilsup{1,3}
    \qquad
    Nils Feldhus$^*$\affilsup{1}
    \\
    \textbf{Simon Ostermann}\affilsup{1,3}
    \qquad
    \textbf{Sebastian M\"oller}\affilsup{1,2}
    \\
    $^1$German Research Center for Artificial Intelligence (DFKI) \\
    $^2$Technische Universit\"at Berlin \\
    $^3$Saarland Informatics Campus\\
    \texttt{\{firstname.lastname\}@dfki.de}
}

\begin{document}
\maketitle

\nnfootnote{* Equally contributed and share the second-authorship.}

\begin{abstract}
Conversational explainable artificial intelligence (ConvXAI) systems based on large language models (LLMs) have garnered significant interest from the research community in natural language processing (NLP) and human-computer interaction (HCI). Such systems can provide answers to user questions about explanations in dialogues, have the potential to enhance users' comprehension and offer more information about the decision-making and generation processes of LLMs. Currently available ConvXAI systems are based on intent recognition rather than free chat, as this has been found to be more precise and reliable in identifying users' intentions. However, the recognition of intents still presents a challenge in the case of ConvXAI, since little training data exist and the domain is highly specific, as there is a broad range of XAI methods to map requests onto. In order to bridge this gap, we present \data{CoXQL}\footnote{\underline{Co}nversational E\underline{x}planation \underline{Q}uery \underline{L}anguage, a word play on \data{CoSQL} \cite{yu-2019-cosql}. Dataset and code are available at \url{https://github.com/DFKI-NLP/CoXQL}.}, the first dataset in the NLP domain for user intent recognition in ConvXAI, covering 31 intents, seven of which require filling multiple slots. 
Subsequently, we enhance an existing parsing approach by incorporating template validations, and conduct an evaluation of several LLMs on \data{CoXQL} using different parsing strategies. We conclude that the improved parsing approach (MP+) surpasses the performance of previous approaches. We also discover that intents with multiple slots remain highly challenging for LLMs.
\end{abstract}
\begin{figure}[t!]
    \centering
    \includegraphics[width=\linewidth]{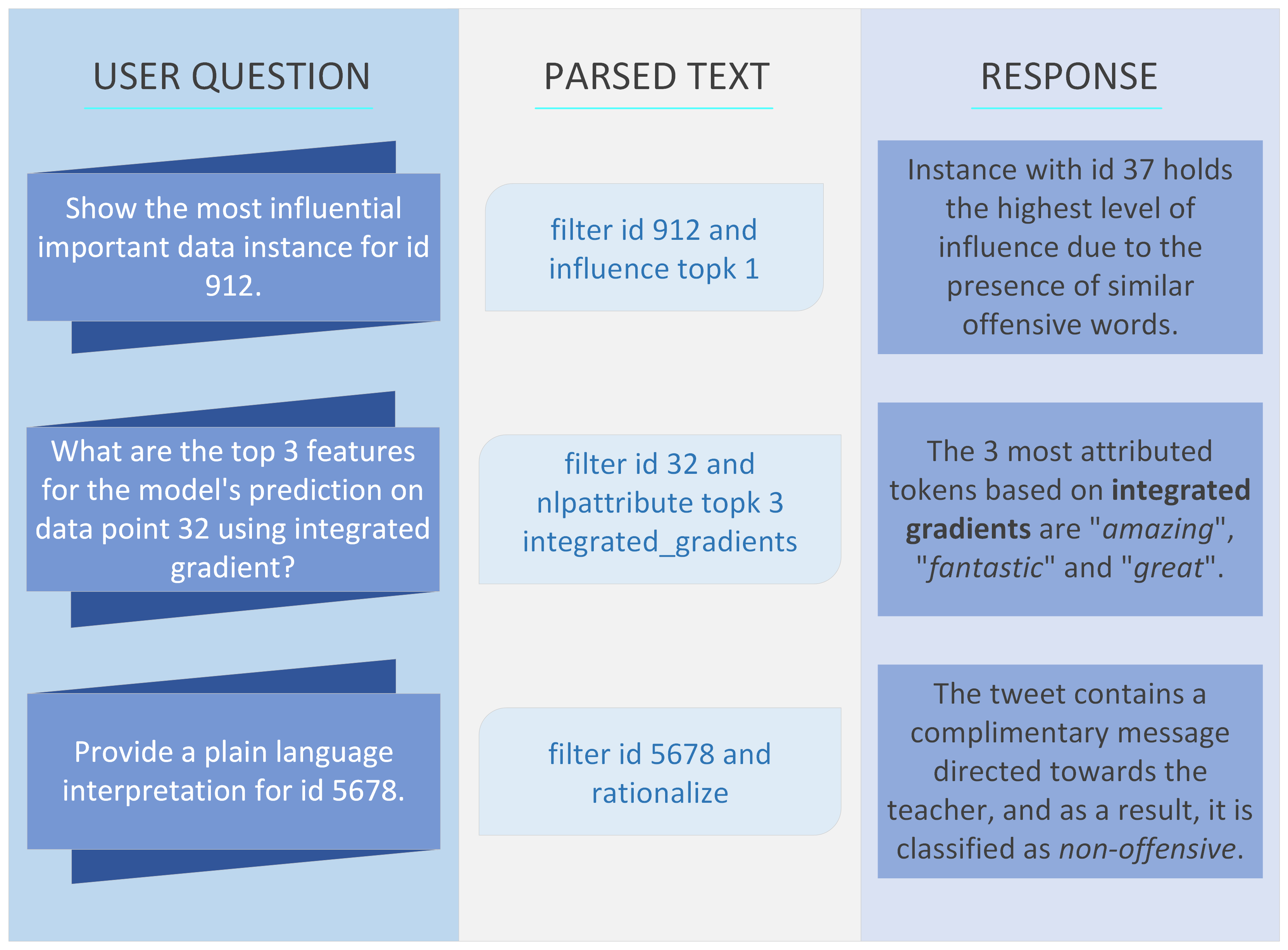}
    \caption{Example utterances consisting of user questions, SQL-like queries (parsed texts) and corresponding responses (not included in \data{CoXQL}) for influence (\texttt{influence}), feature attribution (\texttt{nlpattribute}) and rationalization (\texttt{rationalize}). More examples and operations can be found in Table~\ref{tab:ops} and Table~\ref{tab:example_utterance}.
    }
    \label{fig:examples}
\end{figure}

\begin{table*}[ht!]
    \centering
    \renewcommand*{\arraystretch}{0.92}
    \footnotesize
    \resizebox{\textwidth}{!}{%
        \roundedtable{p{0.2cm}p{5.3cm}|p{10cm}}{

        & \textbf{Operation} 
        & \textbf{Description/Request}
        \\
        
        \toprule
        \multirow{2}{*}{\centering \rotatebox[origin=c]{90}{\tiny{\textbf{Loc.Pr.}}}}
        & \texttt{\textcolor{blue}{predict}(instance) } 
        & Get the prediction for the given instance 
        \\
        
        & \texttt{\textcolor{blue}{likelihood}(instance)} 
        & Calculate the model's confidence (or likelihood) on the given instance 
        \\

        \midrule 
        \multirow{2}{*}{\centering \rotatebox[origin=c]{90}{\tiny{\textbf{Glob.Pr.}}}}
        & \texttt{\textcolor{blue}{mistake}(\{sample|count\}, subset)} 
        & Count or show incorrectly predicted instances
        \\
        
        & \texttt{\textcolor{blue}{score}(subset, metric)} 
        & Determine the relation between predictions and labels
        \\

        \midrule 
        \multirow{2}{*}{\centering \rotatebox[origin=l]{90}{\tiny{\textbf{Loc. Expl.}}}} 
        & \texttt{\textcolor{blue}{nlpattribute}(inst., topk, method) } 
        & Provide feature attribution scores 
        \\
        
        & \texttt{\textcolor{blue}{rationalize}(inst.) } 
        & Explain the output/decision in natural language 
        \\
        
        & \texttt{\textcolor{red}{influence}(inst., topk) } 
        & Provide the most influential training data instances 
        \\

        \midrule 
        \multirow{3}{*}{\centering \rotatebox[origin=c]{90}{\tiny{\textbf{Pertrb.}}}} 
        & \texttt{\textcolor{blue}{cfe}(instance) } 
        & Generate a counterfactual of the given instance 
        \\

        & \texttt{\textcolor{blue}{adversarial}(instance) } 
        & Generate an adversarial example based on the given instance 
        \\
        
        & \texttt{\textcolor{blue}{augment}(instance) } 
        & Generate a new instance based on the given instance 
        \\

        \midrule 
        \multirow{4}{*}{\centering \rotatebox[origin=r]{90}{\tiny{\textbf{Data}}}}
        & \texttt{\textcolor{blue}{show}(instance)} 
        & Show the contents of an instance
        \\
        
        & \texttt{\textcolor{blue}{countdata}(list)} 
        & Count instances
        \\
        
        & \texttt{\textcolor{blue}{label}(dataset)} 
        & Describe the label distribution
        \\
        
        & \texttt{\textcolor{blue}{keywords}(topk)} 
        & Show most common words
        \\
        
        & \texttt{\textcolor{blue}{similar}(instance, topk) } 
        & Show most similar instances
        \\

        \midrule 
        \multirow{3}{*}{\centering \rotatebox[origin=c]{90}{\tiny{\textbf{Mod.}}}} 
        & \texttt{\textcolor{red}{editlabel}(instance)} 
        & Change the true/gold label of a given instance 
        \\
        
        & \texttt{\textcolor{red}{learn}(instance) } 
        & Retrain or fine-tune the model based on a given instance 
        \\
        
        & \texttt{\textcolor{red}{unlearn}(instance) } 
        & Remove or unlearn a given instance from the model 
        \\

        \midrule
        \multirow{6}{*}{\centering \rotatebox[origin=c]{90}{\tiny{\textbf{Meta}}}}
        & \texttt{\textcolor{blue}{function}()} 
        & Explain the functionality of the system
        \\
        
        & \texttt{\textcolor{blue}{tutorial}(op\_name)} 
        & Provide an explanation of the given operation 
        \\
        
        & \texttt{\textcolor{blue}{data}()} 
        & Show the metadata of the dataset
        \\
        
        & \texttt{\textcolor{blue}{model}()} 
        & Show the metadata of the model 
        \\
        
        & \texttt{\textcolor{red}{domain}(query)} 
        & Explain terminology or concepts outside of the system's functionality, but related to the domain
        \\
        }
        }
    \caption{
    Main operations in \data{CoXQL} as they can be requested in a dialogue (Description/Request), mapped onto a partial SQL-like query (Operation) that calls an explanation-generating or data-analyzing method. Red-highlighted operations are currently not implemented in any existing system. Additional logic operations are in Table~\ref{tab:logic_ops}.
    }
    \label{tab:ops}
\end{table*}

\section{Introduction}
There is an increasing number of XAI systems that include user interfaces, facilitating natural language interaction with users \cite{chromik-butz-2021-human-xai-interaction,bertrand-2023-selective}. More recently, there has been a significant development in building ConvXAI systems \cite{lakkaraju-2022-rethinking}, which are guided through intent recognition rather than free-text chatting. The main reason for hard-coding intents is that in a ConvXAI application, there is a need for a maximally faithful conversation, which black-box generation cannot provide \cite{feldhus-2023-interrolang, shen-2023-convxai, wang-etal-2024-llmcheckup}. These systems are designed to answer user questions about explainable language models in dialogues. 
In ConvXAI, intents usually represent the XAI operations supported in the system. The user experience and trust in the system can be negatively impacted when intent recognition fails (e.g., an incorrect mapping of XAI operations can lead to a discrepancy from users' requests). An extensive range of explainability questions has to be processed, which can be formulated in many different ways, depending on the domain of application \cite{lakkaraju-2022-rethinking}. For instance, the user question: \textit{``Clarify id 5678 with a reason.''}, is formulated in different ways but represents the same rationalization intent as depicted in Figure~\ref{fig:examples}.

In this work, 
we present the first dataset for explanation request parsing in the NLP domain, \data{CoXQL} (\S\ref{sec:dataset}). We frame the problem as a text-to-SQL-like task (\S\ref{subsec:text2sql}). \data{CoXQL} consists of user questions and gold parses specifically designed for the XAI domain (Figure~\ref{fig:examples}). 
It can serve as guidance for building ConvXAI systems and as a means to improve explanation intent recognition, where intents are considered as operations supported by ConvXAI systems. 
Moreover, we improve an existing parsing approach based on multi-prompt parsing (MP) \cite{wang-etal-2024-llmcheckup} with additional template checks (\S \ref{subsec:parsing}) and find out that our improved approach (MP+) easily outperforms existing approaches. Lastly, we evaluate several state-of-the-art LLMs with various parsing strategies on \data{CoXQL} for explanation intent recognition (\S\ref{eval:intent}). Our evaluation shows that \data{CoXQL} can be regarded as a benchmark for future research and still presents challenges for state-of-the-art LLMs, especially for accurately recognizing intents (operations) with multiple slots, where slots are finer-grained user preferences regarding XAI operations (e.g., \texttt{topk} and \texttt{integrated gradient} associated with feature attribution in Figure~\ref{fig:examples}).

\section{Related Work}
\label{p:xai_system}
In the majority of previous ConvXAI systems \cite{werner-2020-eric, nguyen-2023-xagent, shen-2023-convxai}, the semantic similarity of sentence embeddings between user query and existing data is used to match the user query with the appropriate operation (Table~\ref{tab:systems_comparison}), known as the nearest neighbor. In contrast, the approach used in \sys{TalkToModel} \cite{slack-2023-talktomodel}, \sys{InterroLang} \cite{feldhus-2023-interrolang} and \sys{LLMCheckup} \cite{wang-etal-2024-llmcheckup} employs LLMs to convert user questions into SQL-like queries (Figure~\ref{fig:examples}). The best performance is achieved in \citet{slack-2023-talktomodel}, \citet{feldhus-2023-interrolang} and \citet{wang-etal-2024-llmcheckup} with a fine-tuned \lm{T5}, an adapter-based \lm{BERT}, and \lm{Llama2} with few-shot prompting, respectively. This parsing approach demonstrates notable enhancements, exceeding a doubling in parsing accuracy compared to the nearest neighbor approach. While they all support no more than 24 operations in their systems, \data{CoXQL} contains in total 31 operations of various complexity ranging from single term operations to operations with multiple slots. The new XAI methods introduced are selected from the literature to help users gain deeper insights into a model's behavior. From our perspective, the current set of operations represents those that a conversational XAI system should be able to handle.



\section{Methodology}
\label{sec:understanding_the_user}

\subsection{Task Framing}
\label{subsec:text2sql}
Building upon the strategy employed by \citet{slack-2023-talktomodel}, \citet{feldhus-2023-interrolang} and \citet{wang-etal-2024-llmcheckup} (\S \ref{p:xai_system}), we treat XAI intent recognition as a text-to-SQL-like task (Figure~\ref{fig:examples}), which can be effectively modeled as a seq2seq task \cite{sutskever-2014-seq2seq}. The generated SQL-like queries should be correctly executable ensuring practical usability and functionality, since failed intent recognition results in incorrect XAI responses, leading to a negative impact on the user experience \cite{feldhus-2023-interrolang}.




\subsection{Supported Operations}
\label{subsec:supported_operation}
We have determined \textbf{23} XAI and supplementary operations, which we show in Table~\ref{tab:ops}, and \textbf{8} additional operations related to logic and filtering depicted in Table~\ref{tab:logic_ops}. The list of available operations (Table~\ref{tab:ops}), including five newly introduced ones (marked in red in Table~\ref{tab:ops}; Appendix~\ref{app:unsupported_operations}), are consolidated from HCI literature \cite{weld-bansal-2019-crafting,liao-2021-question-driven}, the state-of-the-art ConvXAI systems by \citet{slack-2023-talktomodel}, \citet{shen-2023-convxai}, \citet{feldhus-2023-interrolang} and \citet{wang-etal-2024-llmcheckup}, and the taxonomy for LLM interpretation research by \citet{singh-2024-rethinking-interpretability}. Moreover, several operations (Table~\ref{tab:additional_parameters}) are associated with multiple slots, which makes parsing even more challenging for LLMs (Table~\ref{tab:instance_level_error_analysis}). The inclusion of additional fine-grained slots is favored in ConvXAI systems (e.g., \texttt{integrated gradient} in Figure~\ref{fig:examples}), enabling the provision of more informative and multi-faceted explanations \cite{nobani-2021-explainer, wijekoon-2024-tell}.

\subsection{Parsing}
\label{subsec:parsing}

\paragraph{Nearest Neighbor}
Nearest neighbor (NN) relies on comparing semantic similarity between
user query and existing training samples measured by an \lm{SBERT} model\footnote{\url{https://huggingface.co/BAAI/bge-base-en-v1.5}}.
However, as the number of operations and additional slots (e.g., ranges of values, method names) associated with operations grow, the intent recognition accuracy tends to decrease.

\paragraph{Guided Decoding}
Guided Decoding (GD) relies on a predefined grammar to restrict the generated output of LLMs (Figure~\ref{app:gd_grammar}) \cite{shin-2021-constrained}. The parsing prompt used in GD consists of demonstrations that are selected based on their semantic similarity to the desired output (Table~\ref{tab:demonstrations}) \cite{slack-2023-talktomodel}.

\paragraph{Multi-prompt Parsing}
With GD, due to similarity-based pre-selection, the model might miss the demonstrations for the actual operation. Multi-prompt Parsing (MP) \cite{wang-etal-2024-llmcheckup} first queries the model about the main operation by providing coarse-grained demonstrations for all available operations (Table~\ref{tab:ops}) and then selects more fine-grained operation-specific prompts in the next step (Table~\ref{tab:additional_parameters}).

\paragraph{Multi-prompt Parsing with template checking}
Compared to GD, MP is not constrained by the grammar and the parsed text generated by MP is not guaranteed to adhere to the expected template (e.g., the exact order or naming of all slots; Table~\ref{tab:instance_level_error_analysis}). We also find that extracting ids and numerical slots poses a challenge for out-of-the-box prompting with MP. Thus, we improve MP and introduce MP+ that uses additional template checking. This is an important step, since template checking contributes to more reliable parsing that takes both grammar and user input into account\footnote{More details about MP+ are in Appendix~\ref{app:mp_problem}.}. 



\begin{figure}[t!]
    \centering
    \includegraphics[width=\linewidth]{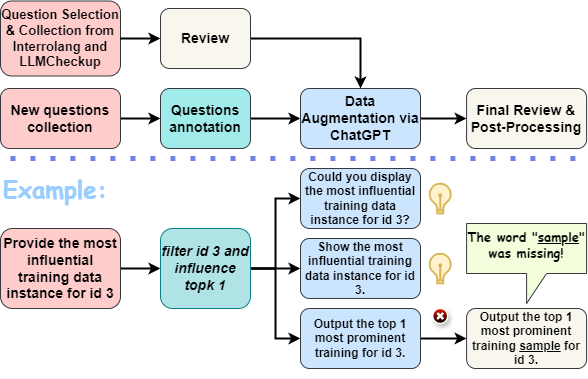}
    \caption{The data collection pipeline of \data{CoXQL}.
    }
    \label{fig:pipeline}
\end{figure}

\section{The \data{CoXQL} Dataset}
\label{sec:dataset}
\subsection{Dataset Creation}
The data creation process of \data{CoXQL} is depicted in Figure~\ref{fig:pipeline}. Based on the predefined set of question and parse pairs from \sys{InterroLang} \cite{feldhus-2023-interrolang} and \sys{LLMCheckup} \cite{wang-etal-2024-llmcheckup}, we selectively choose pairs of question and gold parse for operations marked in blue in Table~\ref{tab:ops}, e.g., by evaluating questions' understandability or topic-parse alignment\footnote{More details are provided in Appendix~\ref{app:data_pipeline}.}. Meanwhile, we manually create new additional pairs for all operations in Table~\ref{tab:ops}, following the way how  questions are raised in \citeposs{feldhus-2023-interrolang} user study. Subsequently, we use \lm{ChatGPT} \cite{ChatGPT} to augment user questions (Figure~\ref{app:prompt_da}) to expand the dataset size. The generated pairs undergo a review process and are post-processed by us if needed (e.g., adding missing words; Figure~\ref{fig:pipeline}).

\begin{figure}[t!]
    \centering
    \includegraphics[width=0.9\linewidth]{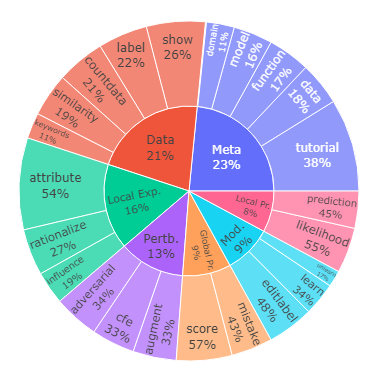}
    \caption{The intent distribution of \data{CoXQL}.
    }
    \label{fig:distribution}
\end{figure}

\subsection{Data Statistics}
After all processing steps, \data{CoXQL} comprises \textbf{1179} pairs of user questions and corresponding SQL-like queries over full SQL parses, 82 of which were post-processed manually. Figure~\ref{fig:distribution} illustrates the intent distribution of \data{CoXQL}. Operations with additional slots in Table~\ref{tab:additional_parameters} have an intentionally higher number of instances compared to others due to their difficulty. Moreover, Table~\ref{tab:example_utterance} provides examples of utterances along with their corresponding parses. Three authors of this work performed the annotations following the annotation instructions as shown in Figure~\ref{fig:annotation_instruction}. We report a token-level inter-annotator agreement of Fleiss' $\kappa$ = 0.87. While LLMs find it challenging to understand different formulations of XAI questions and recognize slots associated with operations simultaneously, these tasks are not as difficult for humans. In addition, we manually crafted \textbf{112} pairs specifically for the test set, which is evaluated in \S \ref{eval:intent}. More details about post-processing and test set are given in Appendix~\ref{app:data_pipeline}.

\section{Evaluation}
\label{eval:intent}

\subsection{Automatic Evaluation}
To assess the ability of interpreting user intents with LLMs, we quantify the performance of seven LLMs\footnote{Two of them, \lm{CodeQWen} \cite{jinze-2023-qwen} and \lm{sqlcoder}, are designed for code and SQL generation. Deployed LLMs are indicated in the left column of Table~\ref{tab:parsing} and in Table~\ref{tab:used_model}.} with different sizes ranging from 1B to 70B, employing four approaches: NN, GD, MP and MP+ (\S \ref{subsec:parsing}) (Table~\ref{tab:parsing}). Performance is calculated by measuring exact match parsing accuracy \cite{talmor-2017-evaluating-semantic-parsing, yu-2018-spider} on \data{CoXQL}. We find that MP falls short of GD on \data{CoXQL} except \lm{CodeQWen1.5} \cite{jinze-2023-qwen}, 
while improved MP (MP+) can easily outperform GD and MP with additional template checks. Among all LLMs and parsing strategies, our findings reveal that \lm{Llama3-70B} with MP+ demonstrates the highest scores, exhibiting a doubling in performance compared to the baseline (NN).

\begin{table}[t!]
    \centering
    \renewcommand*{\arraystretch}{1.0}
    \resizebox{\columnwidth}{!}{%

        \roundedtable{rr|cccc}{
        \textbf{Model} & \textbf{Size} & 
        \colorbox{purple}{NN} & \colorbox{celeste}{GD} & \colorbox{brilliantlavender}{MP} &\colorbox{gblue}{MP+}\\

        \toprule
        Baseline & - & 44.25 & - & - & - \\
        
        \midrule
        \lm{Falcon} & 1B & - & 59.29 & 59.29 & 77.88  \\
        \lm{Pythia} & 2.8B & - & 79.65 & 74.34 & 83.19\\
        \lm{Mistral} & 7B & - & 78.76 & 78.76 & 87.61\\ 
        \lm{Llama3} & 8B & - & 84.07 & 67.26  & 86.73\\
        \lm{Llama3} & 70B & - & 83.19 & 68.14 & 93.81\\

        \midrule
        \lm{CodeQwen1.5} & 7B & - & 65.49 & 67.25 & 85.84 \\
        \lm{sqlcoder} & 7B & - & 86.73 & 79.65 & 88.50\\
        
        }
    }
    \caption{
    Exact match parsing accuracy (in \%) for different models on the \data{CoXQL} test set. \colorbox{purple}{NN} = Nearest Neighbor; \colorbox{celeste}{GD} = Guided Decoding prompted by 20-shots; \colorbox{brilliantlavender}{MP} = Multi-prompt Parsing; \colorbox{gblue}{MP+} = MP with template checks.
    }
    \label{tab:parsing}
\end{table}

\subsection{Error Analysis}

\paragraph{Error analysis at the category level}
Table~\ref{tab:f1} displays $F_1$ scores of each category for different LLMs shown in Table~\ref{tab:used_model}. From Table~\ref{tab:f1}, we find out that GD generally performs better than MP in categories like \texttt{Global Prediction}, \texttt{Local Explanation}, and \texttt{Local Prediction}. MP, however, performs better in categories like \texttt{Data} and \texttt{Modification}. MP+ exceeds the performance of both GD and MP across most categories and models, indicating that the combination of Multi-Prompt parsing with template checks provides a consistent improvement over the individual parsing strategies.

LLMs like \lm{Llama3-8B} and \lm{CodeQWen} benefit the most from the MP+ approach, consistently achieving top scores across multiple categories.
\lm{Falcon} and \lm{Pythia} demonstrate substantial improvements with MP+ over their GD and MP scores, suggesting that MP+ enhances both small-sized and large-sized LMs effectively.

\paragraph{Error analysis at the instance level}
Table~\ref{tab:instance_level_error_analysis} presents parsed texts generated by different LLMs using diverse parsing strategies for the question: \textit{``Top 3 important features for id 3!''}. Tokens in the parsed text that are matched with the gold label are marked with \underline{underlines}. None of the parsed texts match the gold label, regardless of LLMs or parsing strategies, which demonstrates that LLMs still face great challenges when dealing with operations that involve multiple slots. Table~\ref{tab:instance_level_error_analysis} reveals that GD is good in generating top $k$ values accurately, while MP and MP+ tend to correctly generate method names. However, there are instances where MP's generation is incomplete, e.g., the parsed text from \lm{Pythia-2.8B} with MP lacking a numerical value for top $k$. Additionally, GD has a tendency to generate alternative method names like ``lime'' or ``attention'', when the ``default''\footnote{For feature attribution, if no top $k$ value or method is specified, the values \texttt{``all''} and \texttt{``default''} will be used.} should be used when no method name is specified in the users' question (Table~\ref{tab:additional_parameters}). Thus, Table~\ref{tab:instance_level_error_analysis} illustrates that when additional slots are available for operations, LLMs exhibit limitations in fully accurately recognizing every slot (Appendix~\ref{app:parsing}).


\section{Conclusion}
The contributions of this paper are three-fold: Firstly, we present and release the first dataset \data{CoXQL} for explanation request parsing in the NLP domain for ConvXAI systems, featuring 31 intents. Secondly, we improve the previous parsing strategy MP with additional template checks, which considerably improves parsing accuracy. Lastly, we perform a comparative evaluation of seven state-of-the-art LLMs on the \data{CoXQL} data. We find that MP+ outperforms both GD and MP but LLMs still struggle with intents that have multiple slots. In the future, we would like to consider tools like \sys{Langchain}\footnote{\url{https://www.langchain.com/}} to provide more accessible, extensible framework. 


\section*{Limitations}

\data{CoXQL} currently supports only English, and it does not offer multilingual support. However, it is feasible to adapt \data{CoXQL} to target languages through translation.

The complexity of user questions in \data{CoXQL} might be lower when compared to other text-to-SQL datasets that involve complex SQL grammar, such as JOINs, aggregations. Within the current scope, we do not take into account the concatenation of various operations, which could potentially be valuable for users.

All implementations for operations shown in Table~\ref{tab:ops} highlighted in blue can be found in either \sys{TalkToModel} \cite{slack-2023-talktomodel}, \sys{InterroLang} \cite{feldhus-2023-interrolang} or \sys{LLMCheckup} \cite{wang-etal-2024-llmcheckup}. \data{CoXQL} provides annotations for the ones highlighted in red in Table~\ref{tab:ops}. Although none of the existing systems supports additional operations, they can be implemented as described in Appendix~\ref{app:unsupported_operations}. 

While some LLMs, e.g. \lm{Llama3-70B}, can achieve good results in explanation request parsing, their deployment may not always be feasible, e.g., due to resource limitations. This challenge can potentially be addressed by employing active learning techniques on smaller-sized LMs to attain comparable parsing accuracy.

\section*{Acknowledgments}
We thank the reviewers of EMNLP 2024 for their helpful and rigorous feedback.
This work has been supported by the German Federal Ministry of Education and Research as part of the projects XAINES (01IW20005), TRAILS (01IW24005) and VERANDA (16KIS2047).

\bibliography{custom,llmcheckup}

\appendix



\section{Approaches for intent recognition}
Table~\ref{tab:systems_comparison} displays the approaches for intent recognition in the current XAI systems.

\begin{table}[h!]
    \centering
    \renewcommand*{\arraystretch}{1.0}
    \resizebox{\columnwidth}{!}{%

    \begin{tabular}{|r|c|cc|c|}

    \toprule
    \textbf{XAI System}
    & \multicolumn{3}{c|}{\textbf{Intent recognition}}
    & \multirow{2}{0.9cm}{\small \textbf{\textls[-50]{Text-to-SQL}}}
    \\

    \textbf{Implementations}
    & Embeds & Fine-Tuned & Few-Shot 
    & 
    \\

    \midrule


    \citet{werner-2020-eric}
    & fastText & &
    & 
    \\

    \citet{torri-2021-textual-explanations}
    & & \lm{GPT-2} &
    & 
    \\

    \citet{slack-2023-talktomodel}
    & \lm{MPNet} & \lm{T5} & \lm{GPT-J}
    & \present
    \\

    \textls[-50]{\citet{nguyen-2023-xagent}}
    & \lm{SimCSE} & &
    & 
    \\


    \citet{shen-2023-convxai}
    & \lm{SciBERT} & &
    & 
    \\

    \multirow{2}{*}{\textls[-50]{\citet{feldhus-2023-interrolang}}}
    & \multirow{2}{*}{\lm{MPNet}} & \textls[-50]{\lm{BERT+Adap}}, & \multirow{2}{*}{\lm{GPT-Neo}}
    & \multirow{2}{*}{\present}
    \\

    & 
    & \lm{FLAN-T5}
    & 
    & 
    \\

    \citet{wang-etal-2024-llmcheckup}
    & \lm{MPNet}
    & & \lm{Llama2}
    & \present \\

    Ours
    & \lm{bge-base}
    & & \lm{Llama3}
    & \present
    \\
    
    \bottomrule
    
    \end{tabular}
    }
    \caption{Approaches for intent recognition in conversational XAI systems using LM embeddings, fine-tuned LMs and LLMs with few-shot prompting.
    }
    \label{tab:systems_comparison}
\end{table}

\section{Guided decoding}
\subsection{Example grammar}
\begin{figure*}[t!]
    \centering
    \includegraphics[width=\linewidth]{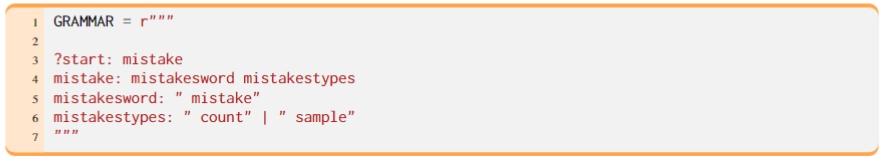}
    \caption{Example grammar of \texttt{mistake} operation with additional slots ``count'' and ``sample''.
    }
    \label{app:gd_grammar}
\end{figure*}

Figure~\ref{app:gd_grammar} shows the grammar for \texttt{mistake} operation with additional slots \texttt{count} and \texttt{sample}.

\begin{table*}[t!]
    \centering
    \renewcommand*{\arraystretch}{1.0}
    \resizebox{\textwidth}{!}{%
        \roundedtable{c|c}{

        \textbf{Type} & \textbf{Text} \\
        \toprule
        User question & Can you show me how much data the model predicts incorrectly?\\
        \midrule
        \multirow{3}{*}{Selected demonstration} & Tell me the amount of data the model predicts falsely.\\
        & Can you demonstrate how many data points are predicted wrongly?\\
        & Show me some data you predict incorrectly. \\

        }
        }
    \caption{
    Selected top 3 demonstrations based on semantic similarity.
    }
    \label{tab:demonstrations}
\end{table*}

\subsection{Demonstration selection}
As described in \S \ref{subsec:parsing}, for guided decoding, the parsing prompt will contain demonstrations which are selected based semantic similarity. Table~\ref{tab:demonstrations} shows the top 3 similar selected demonstrations for the user question ``Can you show me how much data the model predicts incorrectly?''.

\section{Example utterance from \data{CoXQL}}
Table~\ref{tab:example_utterance} provides example utterances corresponding to each operation listed in Table~\ref{tab:ops}.

\begin{table*}[ht!]
    \centering
    \footnotesize
    \resizebox{\textwidth}{!}{%
        \begin{tabular}{|p{0.2cm}p{1.75cm}|p{8cm}|p{6cm}|}

        \toprule
        & \textbf{Intent class} & \textbf{Example utterance} & \textbf{Gold parse}
        \\
        
        \toprule
        \multirow{2}{*}{\centering \rotatebox[origin=c]{90}{\tiny{\textbf{Loc.Pr.}}}}
        & \texttt{\textcolor{blue}{predict} } 
        & What is the prediction for data point number 9130? & \texttt{filter id 9130 and predict}
        \\
        
        & \texttt{\textcolor{blue}{likelihood}} 
        & Give me the confidence score for this prediction on id 15? & \texttt{filter id 15 and likelihood}
        \\

        \midrule 
        \multirow{2}{*}{\centering \rotatebox[origin=c]{90}{\tiny{\textbf{Glob.Pr.}}}}
        & \texttt{\textcolor{blue}{mistake}} 
        & Tell me the amount of data the model predicts falsely. & \texttt{mistake count}
        \\
        
        & \texttt{\textcolor{blue}{score}} 
        & Give me the accuracy on the data.
        & \texttt{score accuracy}
        \\

        \midrule 
        \multirow{2}{*}{\centering \rotatebox[origin=c]{90}{\tiny{\textbf{Loc. Exp.}}}} 
        & \texttt{\textcolor{blue}{attribute}} 
        & Why do you predict instance 2451? & \texttt{filter id 2451 and nlpattribute default}
        \\
        
        & \texttt{\textcolor{blue}{rationalize}} 
        & Generate a natural language explanation for id 2222. & \texttt{filter id 2222 and rationalize}
        \\
        
        & \texttt{\textcolor{red}{influence}} 
        & Show the most influential important data instance for id 912. & \texttt{filter id 912 and influence topk 1}
        \\

        \midrule 
        \multirow{1}{*}{\centering \rotatebox[origin=r]{90}{\tiny{\textbf{Pertrb.}}}} 
        & \texttt{\textcolor{blue}{cfe}} 
        & How would you flip the prediction for id 23? & \texttt{filter id 23 and cfe}
        \\

        & \texttt{\textcolor{blue}{adversarial}} 
        & How would you construct an adversarial example for the model's prediction on id 23? & \texttt{filter id 23 and adversarial}
        \\
        
        & \texttt{\textcolor{blue}{augment}} 
        & Can you modify and generate a new instance from id 100? & \texttt{filter id 100 and augment}
        \\

        \midrule 
        \multirow{2}{*}{\centering \rotatebox[origin=r]{90}{\tiny{\textbf{Data}}}}
        & \texttt{\textcolor{blue}{show}} 
        & Could you show me data point number 215?
        & \texttt{filter id 215 and show} \\
        
        & \texttt{\textcolor{blue}{countdata}} 
        & Count the total number of data points. & \texttt{countdata}
        \\
        
        & \texttt{\textcolor{blue}{label}} 
        & Please show what the gold labels are.
        & \texttt{label}
        \\
        
        & \texttt{\textcolor{blue}{keywords}} 
        & What are the most frequent keywords in the data?
        & \texttt{keywords topk 1}
        \\
        
        & \texttt{\textcolor{blue}{similar}} 
        & Is it possible to retrieve an example that is similar to id 12?
        & \texttt{filter id 12 and similarity topk 1}
        \\

        \midrule 
        \multirow{2}{*}{\centering \rotatebox[origin=r]{90}{\tiny{\textbf{Mod.}}}} 
        & \texttt{\textcolor{red}{editlabel}} 
        & Edit the label of id 2894 to the specified label.
        & \texttt{filter id 2894 and editlabel}
        \\
        
        & \texttt{\textcolor{red}{learn}} 
        & Apply training to the model using instance 473.
        & \texttt{filter id 473 and learn}
        \\
        
        & \texttt{\textcolor{red}{unlearn}} 
        & Can you unlearn id 530 from the model?
        & \texttt{filter id 530 and unlearn}
        \\

        \midrule
        \multirow{2}{*}{\centering \rotatebox[origin=r]{90}{\tiny{\textbf{Meta}}}}
        & \texttt{\textcolor{blue}{function}} 
        & Tell me a bit more about what I can do here.
        & \texttt{function}
        \\
        
        & \texttt{\textcolor{blue}{tutorial}} 
        & What's data augmentation?
        & \texttt{qatutorial qada}
        \\
        
        & \texttt{\textcolor{blue}{data}} 
        & Tell me a bit more about the data please.
        & \texttt{data}
        \\
        
        & \texttt{\textcolor{blue}{model}} 
        & It would be very useful if you could provide a description of the model!
        & \texttt{model}
        \\
        
        & \texttt{\textcolor{red}{domain}} 
        & Can you clarify terms or concepts that are relevant to the domain but not directly related to the system's functionality?
        & \texttt{domain}
        \\
        \bottomrule
        \end{tabular}
        }
    \caption{
     Intent classes, example utterance from \data{CoXQL} and corresponding gold parse.
    }
    \label{tab:example_utterance}
\end{table*}

\section{Additional slots for operations}
\label{app:additional_parameters}
Table~\ref{tab:additional_parameters} shows operations with additional slots.
\begin{table*}[t!]
    \centering
    \renewcommand*{\arraystretch}{1.0}
    \resizebox{\textwidth}{!}{%
        \roundedtable{c|c|c}{

        \textbf{Operation} & \textbf{Additional Slots} & \textbf{\#Additional Slots}\\
        \toprule
        \texttt{influence} & $topk$ & 1\\
        \midrule
        \texttt{keywords} & $topk$ & 1\\
        \midrule
        \texttt{similarity} & $topk$ & 1\\
        \midrule
        \texttt{mistake} & sample, count & 2\\
        \midrule
        \texttt{score} & accuracy, precision, recall, $f_1$, roc & 5\\

        \midrule
        
        \multirow{2}{*}{\texttt{attribute}} & all, $topk$, default & \multirow{2}{*}{7}\\ & attention, lime, integrated gradient, inputxgradient & \\

        \midrule
        
        \multirow{2}{*}{\texttt{tutorial}} & qaattribute, qarationalize, qainfluence, qacfe & \multirow{2}{*}{9}\\ & qaadversarial, qaaugment, qaeditlabel, qalearn, qaunlearn & \\

        }
        }
    \caption{
    Additional slots for operations.
    }
    \label{tab:additional_parameters}
\end{table*}

\section{Filter and logic operations}
\label{app:logic}
In addition to the operations displayed in Table~\ref{tab:ops}, we have also incorporated operations related to logic and filtering, as depicted in Table~\ref{tab:logic_ops}. While \sys{InterroLang} \cite{feldhus-2023-interrolang} and \sys{LLMCheckup} \cite{wang-etal-2024-llmcheckup} already include \texttt{predfilter}, \texttt{labelfilter} and \texttt{previousfilter}, we introduce a new filter called \texttt{lengthfilter}, which allows for dataset filtering based on the length of the instances at various levels of granularity, such as character, token, or sentence.

Those aforementioned filters allows for a wide range of possibilities in analyzing and manipulating the dataset based on various conditions and interests. For instance, one can examine data points where the predicted label differs from the golden label using a combination of \texttt{labelfilter} and \texttt{predfilter}. In addition, all filters can be interconnected with operations listed in Table~\ref{tab:ops}.

\begin{table*}[ht!]
    \small
    \centering
    \renewcommand*{\arraystretch}{1}
    \footnotesize
    \resizebox{\textwidth}{!}{%
        \begin{tabular}{|p{0.2cm}p{4.2cm}|p{6.3cm}|p{3.8cm}|}

        \toprule
        & \textbf{Operation} 
        & \multicolumn{2}{l|}{\textbf{Description/Request}}
        
        \\

        \toprule
        \multirow{2}{*}{\centering \rotatebox[origin=c]{90}{\textbf{Filter}}}
        & \texttt{\textcolor{blue}{filter}(id)}
        & \multicolumn{2}{l|}{Access single instance by its ID} 
        \\
        
        & \texttt{\textcolor{blue}{predfilter}(label)}
        & \multicolumn{2}{l|}{Filter the dataset according to the model's predicted label}
        \\
        
        & \texttt{\textcolor{blue}{labelfilter}(label)}
        & \multicolumn{2}{l|}{Filter the dataset according to the true/gold label given by the dataset}
        \\
        
        & \texttt{\textcolor{red}{lengthfilter}(level, len)}
        & \multicolumn{2}{l|}{Filter the dataset by length of the instance (characters, tokens, …)}
        \\
        
        & \texttt{\textcolor{blue}{previousfilter}()}
        & \multicolumn{2}{l|}{Filter the dataset according to outcome of previous operation}
        \\
        
        & \texttt{\textcolor{blue}{includes}(token)}
        & \multicolumn{2}{l|}{Filter the dataset by token occurrence}
        \\

        \midrule 
        \multirow{1}{*}{\centering \rotatebox[origin=c]{90}{\textls[-50]{\textbf{Logic}}}}
        & \texttt{\textcolor{blue}{and}(op1, op2)} 
        & \multicolumn{2}{l|}{Concatenate multiple operations} 
        \\
        
        & \texttt{\textcolor{blue}{or}(op1, op2)} 
        & \multicolumn{2}{l|}{Select multiple filters} 
        \\
        \bottomrule
        \end{tabular}
        }
    \caption{
    Additional logic operations in \data{CoXQL}.
    }
    \label{tab:logic_ops}
\end{table*}

\section{Multi-prompt parsing}
\label{app:mp_problem}
As indicated in Section~\ref{subsec:parsing}, MP is not constraint by the predefined grammar. From Table~\ref{tab:f1}, we found that extracting ids and numerical slots poses a significant challenge for out-of-the-box prompting, especially for those LLMs that have less parameters (e.g., \lm{falcon-1B} or \lm{Pythia-2.8B}). Vanilla MP shows lower performance on operations from Table~\ref{tab:additional_parameters} that require several slots  (e.g., \texttt{Global Prediction} and \texttt{Local Explanation}, see Table~\ref{tab:f1}). 
The lower performance of MP compared to GD can be attributed to the fact that MP tends to generate a larger volume of tokens/slots, given MP's lack of constraints imposed by grammar. For instance, in the case of \texttt{score} operation, which can take values such as \texttt{accuracy}, \texttt{precision}, \texttt{roc}, \texttt{recall}, or \texttt{f1} as additional slots, MP has a tendency to produce more than one metric name. 
Thus, we propose MP+, which applies additional template checks on the generated parsed text and can achieve best performance compared to GD and MP (\S\ref{eval:intent}).

\section{Prompt design}
\label{app:prompt_design}
\begin{figure*}[t!]
    \centering
    \includegraphics[width=\linewidth]{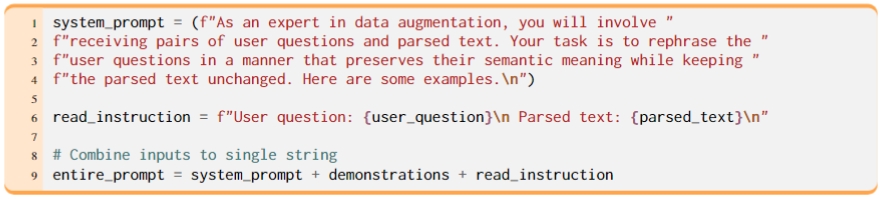}
    \caption{Simplified version of the Python code showing the data augmentation prompt using \lm{ChatGPT} to generate additional data points for \data{CoXQL}.
    }
    \label{app:prompt_da}
\end{figure*}

Figure~\ref{app:prompt_da} shows the prompt used with \lm{ChatGPT} to produce additional data points for \data{CoXQL}.

\section{Annotation Instructions}
\begin{figure}[t!]
    \centering
    \includegraphics[width=0.9\linewidth]{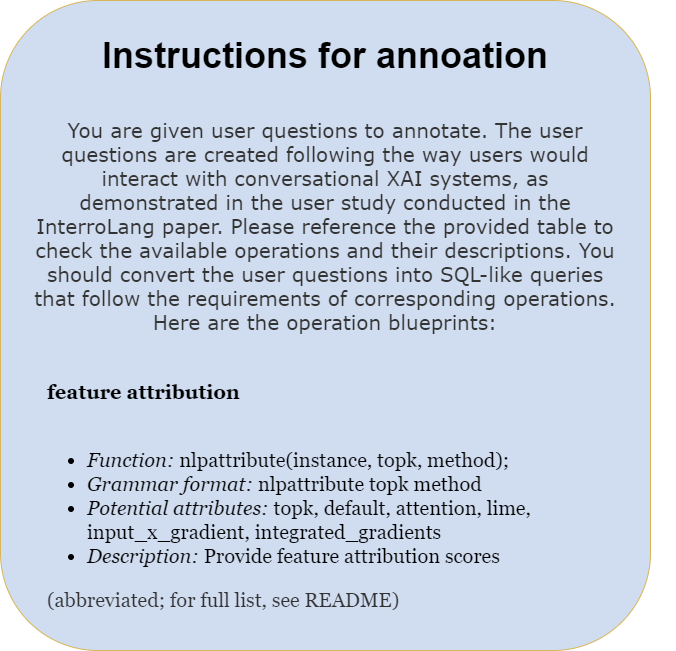}
    \caption{The annotation instructions for \data{CoXQL}.
    }
    \label{fig:annotation_instruction}
\end{figure}

Figure~\ref{fig:annotation_instruction} displays the annotation instructions for \data{CoXQL}.

\section{Data collection}
\label{app:data_pipeline}
\paragraph{Data collection pipeline} We employ a selective approach where we choose question and parse pairs from \sys{InterroLang} \cite{feldhus-2023-interrolang} and \sys{LLMCheckup} \cite{wang-etal-2024-llmcheckup} specifically for operations that are also present in \data{CoXQL}. Subsequently, we thoroughly review all the collected user questions, assessing aspects such as readability, understandability, and coherence. Additionally, we ensure that the purpose or topic conveyed within the user question aligns with the corresponding parse. If post-processing is required, such as in the case of pairs from \sys{InterroLang} \cite{feldhus-2023-interrolang} and \sys{LLMCheckup} \cite{wang-etal-2024-llmcheckup}, and pairs generated by \lm{ChatGPT}, we may need to reformulate the user questions or potentially modify the parsed text based on the intended meaning or intent of the questions. E.g. when we use \lm{ChatGPT} to augment the user question \textit{``Why do you predict instance id 31 using input gradient?''}, which should be parsed as ``\texttt{filter id 31 and nlpattribute \underline{all} input\_x\_gradient}''. Since \texttt{nlpattribute} operation (feature attribution) has many additional slots (Table~\ref{tab:additional_parameters}), \lm{ChatGPT} generates the parsed text of the mentioned question as ``\texttt{filter id 31 and nlpattribute \underline{topk 1} input\_x\_gradient}'' (the additional slot should be \texttt{all} instead of \texttt{topk 1} because user question does not specify the top $k$ values and thus \texttt{all} should be set as default), although we instruct \lm{ChatGPT} to not change the parsed text in the prompt (Figure~\ref{app:prompt_da}). In such a case, we have to post-process the parsed text by changing ``topk 1'' to ``all''.

\paragraph{Test set creation} 
\citet{feldhus-2023-interrolang} conducts a user study to evaluate the quality of explanations generated by \sys{InterroLang}. The user questions, along with their corresponding answers and parsed texts from this user study, are publicly accessible\footnote{\url{https://github.com/DFKI-NLP/InterroLang/blob/main/feedback}}. Inspired by \citeposs{feldhus-2023-interrolang} approach, we adopt a similar strategy and a subset of the test set is created following the way how  questions are raised from the user study.

\section{Operations not supported in current XAI dialogue systems}
\label{app:unsupported_operations}
We introduce five new operations, which are currently not present in the existing ConvXAI systems outlined in Table~\ref{tab:ops} and Table~\ref{tab:logic_ops} marked in red. \texttt{influence} operation enables the retrieval of the most influential training data contributing to the result \cite{han-2020-explaining}. \texttt{editlabel} operation allows for the modification of the golden label for a specific instance. With the \texttt{learn} and \texttt{unlearn} operations, the deployed model can be additionally fine-tuned with or without a particular instance. The \texttt{domain} operation provides information regarding terminology or concepts relevant to our domain but not covered by the system.

We outline here how we would implement them:
\begin{itemize}[noitemsep,topsep=0pt]
    \item \texttt{influence(instance, topk)}: To calculate influential training instances, \sys{Captum} provides a tutorial for the TracIn method: \url{https://captum.ai/tutorials/TracInCP_Tutorial}. However, it is quite expensive to execute on LLMs.
    \item Modification operations are related to explanatory debugging, an area of research surveyed in \citet{lertvittayakumjorn-toni-2021-explanation-based-human-debugging}. A representative system is \sys{XMD} \cite{lee-2023-xmd}.
    \item \texttt{domain(query}): The entire user question is provided to the LLM and the operation is treated as an open-domain question answering task similar to the \texttt{rationalize} operation.
    \item \texttt{lengthfilter(level, len)} is straightforward to implement and only considers the dataset instances with a length above or below some character, token, word, or sentence count (specified by the granularity \texttt{level} slot).
\end{itemize}

Additionally, we want to point out that in practical applications of XAI systems, it is common to encounter a significant number of questions belonging to \texttt{domain} operation. In such cases, the \sys{Toolformer} \cite{schick-2023-toolformer} can be integrated and utilized to directly access relevant tools or APIs associated with the domain-specific questions. 

\data{CoXQL} deliberately excluded attention head and circuit analyses which are not well-suited for conversational explanations and are dependant on visualization rather than text as a modality for explanation. We propose to use dedicated tools for those purposes \cite{tufanov-2024-lm-transparency-tool}.

\section{Parsing accuracy evaluation}
\label{app:parsing}
\subsection{Models for parsing accuracy evaluation}
\begin{table*}[t!]
    \centering
    \resizebox{\textwidth}{!}{%
        \begin{tabular}{rccc}

        \toprule
        \textbf{Name}& \textbf{Citation} & \textbf{Size} & \textbf{Link}\\

        \midrule
        \lm{Falcon} & \citet{penedo-2023-refinedweb} & 1B & \url{https://huggingface.co/tiiuae/falcon-rw-1b}\\
        \lm{Pythia} & \citet{biderman-2023-pythia} & 2.8B & \url{https://huggingface.co/EleutherAI/pythia-2.8b-v0}\\
        \lm{Mistral} & \citet{jiang-2023-mistral} & 7B & \url{https://huggingface.co/mistralai/Mistral-7B-v0.1} \\
        \lm{CodeQwen1.5} & \citet{jinze-2023-qwen} & 7B & \url{https://huggingface.co/Qwen/CodeQwen1.5-7B-Chat}\\
        \lm{sqlcoder} & $\text{n.a.}^*$ & 7B & \url{https://huggingface.co/defog/sqlcoder-7b-2}\\ 
        \lm{Llama 3} & $\text{n.a.}^*$ & 8B & \url{https://huggingface.co/meta-llama/Meta-Llama-3-8B} \\
        \lm{Llama 3} & $\text{n.a.}^*$ & 70B & \url{https://huggingface.co/meta-llama/Meta-Llama-3-70B}\\
        
        \bottomrule
        \end{tabular}
        }
    \caption{
    Deployed LMs for parsing accuracy evaluation. *No paper published, with \lm{GitHub} link only: \url{https://github.com/meta-llama/llama3} and \url{https://github.com/defog-ai/sqlcoder}. 
    }
    \label{tab:used_model}
\end{table*}
Table~\ref{tab:used_model} lists all LLMs that are evaluated for parsing. We used A100 and H100 for parsing accuracy evaluation, which is done within 1 hour per setting.

\subsection{Error analysis}
\label{subsec:error_analysis}

\begin{table*}[t!]
    \centering
    \resizebox{\textwidth}{!}{%
        \begin{tabular}{@{\extracolsep{\fill}}cc|ccccc|cc}

        \toprule
         \textbf{Category} & \textbf{Strat.}  & \textls[-50]{\lm{Falcon}} & \textls[-50]{\lm{Pythia}} & \textls[-50]{\lm{Mistral}} & \textls[-50]{\lm{Llama3-8B}} & \textls[-50]{\lm{Llama3-70B}} & \lm{CodeQWen} & \lm{sqlcoder} \\
        
        \midrule
        \textbf{Data} & \colorbox{celeste}{GD} & 63.43 & 89.77 & 71.88 & 91.67 & 85.42 & 77.08 & 80.21\\
        
        \textbf{Glb. Pr.} & \colorbox{celeste}{GD} & 72.97 & 93.14 & 93.14 & 100.00 & 100.00 & 83.33 & 100.00\\
        
        \textbf{Loc. Ex.} & \colorbox{celeste}{GD} & 53.85 & 80.77 & 80.77 & 84.62 & 84.62 & 73.08 & 84.62\\
        
        \textbf{Loc. Pr.} & \colorbox{celeste}{GD} & 66.67 & 100.00 & 100.00 & 100.00 & 100.00 & 66.67 & 100.00\\
        
        \textbf{Meta} & \colorbox{celeste}{GD} & 70.04  & 64.05 & 75.00 & 69.15 & 75.75 & 54.54 & 85.71\\
        
        \textbf{Modi.} & \colorbox{celeste}{GD} & 36.36 & 63.64 & 54.55 & 63.64 & 63.64 & 54.55 & 72.73\\
        
        \textbf{Pert.} & \colorbox{celeste}{GD} & 60.00 & 100.00 & 100.00 & 100.00 & 100.00 & 70.00 & 100.00\\

        \midrule

        \textbf{Data} & \colorbox{brilliantlavender}{MP} & 65.63 & 70.83 & 91.67 & 81.02 & 85.02 & 82.67 & 100.00 \\
        
        \textbf{Glb. Pr.} & \colorbox{brilliantlavender}{MP} & 0.00 & 0.00 & 29.33 & 54.86 & 8.00 & 32.00 & 93.33\\
        
        \textbf{Loc. Ex.} & \colorbox{brilliantlavender}{MP} & 26.92 & 11.54 & 46.15 & 51.65 & 30.77 & 26.92  & 61.53\\
        
        \textbf{Loc. Pr.} & \colorbox{brilliantlavender}{MP} & 44.44 & 92.59 & 81.48 & 70.37 & 70.37 & 55.56 & 81.48\\
        
        \textbf{Meta} & \colorbox{brilliantlavender}{MP} & 85.02  & 88.89 & 79.05 & 67.70 & 96.77 & 76.94 & 80.56\\
        
        \textbf{Modi.} & \colorbox{brilliantlavender}{MP} & 63.63 & 81.82 & 90.91 & 81.82 & 72.73 & 90.91 & 81.82\\
        
        \textbf{Pert.} & \colorbox{brilliantlavender}{MP} & 100.00 & 100.00 & 100.00 & 60.00 & 70.00 & 90.00 & 50.00\\

        \midrule

        \textbf{Data} & \colorbox{gblue}{MP+} & 73.96 & 91.67 & 100.00 & 95.83 & 95.19 & 97.50 & 100.00\\
        
        \textbf{Glb. Pr.} & \colorbox{gblue}{MP+} & 69.45 & 68.14 & 80.55 & 86.77 & 91.11 & 84.55 & 89.63\\
        
        \textbf{Loc. Ex.} & \colorbox{gblue}{MP+} & 70.94 & 58.65 & 72.22 & 85.04 & 87.18 & 76.07 & 74.79\\
        
        \textbf{Loc. Pr.} & \colorbox{gblue}{MP+} & 44.44 & 100.00 & 81.48 & 70.37 & 100.00 & 66.67 & 88.89\\
        
        \textbf{Meta} & \colorbox{gblue}{MP+} & 87.40 & 88.89 & 82.94 & 72.78 & 93.23 & 78.71 & 82.24\\
        
        \textbf{Modi.} & \colorbox{gblue}{MP+} & 90.91 & 90.91 & 100.00 & 100.00 & 100.00 & 100.00 & 100.00\\
        
        \textbf{Pert.} & \colorbox{gblue}{MP+} & 100.00 & 90.00 & 100.00 & 100.00 & 100.00 & 100.00 & 90.00\\
        
        \bottomrule
        \end{tabular}
    }
    \caption{
    $F_1$ scores of each category for different LMs on \data{CoXQL} test set. \colorbox{celeste}{GD} = Guided Decoding prompted by 20-shots; \colorbox{brilliantlavender}{MP} = Multi-Prompt parsing; \colorbox{gblue}{MP+} = MP with template checks. 
    }
    \label{tab:f1}
\end{table*}

\begin{table*}[t!]
    \centering
    \renewcommand*{\arraystretch}{1}
    \resizebox{\textwidth}{!}{%
        \begin{tabular}{|c|c|c|c|}

        \toprule
        \textbf{Model} & \textbf{Strategy} & \textbf{Parsed Text} & \textbf{Correctness}\\
        
        \midrule
        \multirow{3}{*}{\lm{Falcon-1B}} & \colorbox{celeste}{GD} &  \texttt{filter id 3 and nlpattribute \underline{topk 3} lime} & \ding{55}\\
        & \colorbox{brilliantlavender}{MP} & \texttt{filter id 3 and nlpattribute attention all}& \ding{55}\\
        & \colorbox{gblue}{MP+} & \texttt{filter id 3 and nlpattribute all \underline{default}}& \ding{55}\\
        \hline
        
        \multirow{3}{*}{\lm{Pythia-2.8B}} & \colorbox{celeste}{GD} &   \texttt{filter id 3 and nlpattribute \underline{topk 3} lime}& \ding{55}\\
        & \colorbox{brilliantlavender}{MP} & \texttt{filter id 3 and nlpattribute input\_x\_gradient \underline{topk}}& \ding{55}\\
        & \colorbox{gblue}{MP+} & \texttt{filter id 3 and nlpattribute all \underline{default}}& \ding{55}\\
        \hline
        
        \multirow{3}{*}{\lm{Mistral-7B}} & \colorbox{celeste}{GD} &  \texttt{filter id 3 and nlpattribute \underline{topk 3} lime} & \ding{55}\\
        & \colorbox{brilliantlavender}{MP} & \texttt{filter id 3 and nlpattribute all \underline{default}}& \ding{55}\\
        & \colorbox{gblue}{MP+} & \texttt{filter id 3 and nlpattribute all \underline{default}}& \ding{55}\\
        \hline

        \multirow{3}{*}{\lm{CodeQwen1.5-7B}}  & \colorbox{celeste}{GD} &   \texttt{filter id 3 and nlpattribute \underline{topk 3} attention}& \ding{55}\\
        & \colorbox{brilliantlavender}{MP} & \texttt{filter id 3 and nlpattribute all \underline{default}}& \ding{55}\\
        & \colorbox{gblue}{MP+} & \texttt{filter id 3 and nlpattribute all \underline{default}}& \ding{55}\\
        \hline
        
        \multirow{3}{*}{\lm{sqlcoder-7B}} & \colorbox{celeste}{GD} &  \texttt{filter id 3 and nlpattribute \underline{topk 3} lime} & \ding{55}\\
        & \colorbox{brilliantlavender}{MP} & \texttt{filter id 3 and nlpattribute all \underline{default}}& \ding{55}\\
        & \colorbox{gblue}{MP+} & \texttt{filter id 3 and nlpattribute all \underline{default}}& \ding{55}\\
        \hline
        
        \multirow{3}{*}{\lm{Llama3-8B}} & \colorbox{celeste}{GD} &  \texttt{filter id 3 and nlpattribute \underline{topk 3} attention} & \ding{55}\\
        & \colorbox{brilliantlavender}{MP} & \texttt{filter id 3 and nlpattribute all \underline{default}}& \ding{55}\\
        & \colorbox{gblue}{MP+} & \texttt{filter id 3 and nlpattribute all \underline{default}}& \ding{55}\\
        \hline
        
        \multirow{3}{*}{\lm{Llama3-70B}}  & \colorbox{celeste}{GD} &  \texttt{filter id 3 and nlpattribute \underline{topk 3} attention} & \ding{55}\\
        & \colorbox{brilliantlavender}{MP} & \texttt{filter id 3 and nlpattribute \underline{topk} all}& \ding{55}\\
        & \colorbox{gblue}{MP+} & \texttt{filter id 3 and nlpattribute all \underline{default}}& \ding{55}\\

        \bottomrule
        \end{tabular}
    }
    \caption{ Parsed texts generated by various LMs employing different parsing strategies for the user question: \textit{``Top 3 important features for id 3!''}, where the gold label is
    \texttt{filter id 3 and nlpattribute \underline{topk 3 default}}. Tokens associated with additional attributes that are matched with the gold label are marked with \underline{underlines}. \ding{55} marks a parsed text that does not match the gold label. \colorbox{celeste}{GD} = Guided Decoding prompted by 20-shots; \colorbox{brilliantlavender}{MP} = Multi-prompt Parsing; \colorbox{gblue}{MP+} = MP with template checks.
    }
    \label{tab:instance_level_error_analysis}
\end{table*}

A detailed error analysis for each category is given in Table~\ref{tab:f1}. GD outperforms MP when operations involve a greater number of additional slots (Table~\ref{tab:additional_parameters}), which is due to MP's tendency to generate a higher volume of slots and MP not being constrained by grammar. 
However, MP+ can achieve overall better results. Additionally, Table~\ref{tab:instance_level_error_analysis} shows the parsed texts of the question: \textit{``Top 3 important features for ID 3!''}, generated by all deployed LLMs. None of them can fully match the gold parse, regardless of LLMs or parsing strategies, which demonstrates that LLMs still face great challenges when dealing with operations that involve multiple slots.
\end{document}